\pdfoutput=1

\documentclass[11pt]{article}

\usepackage{ACL2023}
\usepackage{multirow}
\usepackage{times}
\usepackage{latexsym}
\usepackage{graphicx}
\usepackage{color-edits}
\usepackage{amsmath}
\usepackage{booktabs}
\usepackage{colortbl}
\usepackage{enumitem}

\addauthor[Sameer]{sj}{red}

\usepackage[T1]{fontenc}

\usepackage[utf8]{inputenc}

\usepackage{microtype}

\usepackage{inconsolata}

\newenvironment{itemize*}%
 {\leftmargini=10pt\begin{itemize}%
  \setlength{\itemsep}{0pt}%
  \setlength{\parskip}{0pt}%
  }%
 {\end{itemize}}
\newenvironment{enumerate*}%
 {\begin{enumerate}%
  \setlength{\itemsep}{0pt}%
  \setlength{\parskip}{0pt}}%
 {\end{enumerate}}

\title{Multi-Dimensional Evaluation of Text Summarization with \\In-Context Learning}

\author{
\textbf{Sameer Jain}$^{1}$  \quad
\textbf{Vaishakh Keshava}$^{1}$  \quad
\textbf{Swarnashree Mysore Sathyendra}$^{1}$ \\
\textbf{Patrick Fernandes}$^{1, 2}$  \quad
\textbf{Pengfei Liu}$^{1}$ \quad
\textbf{Graham Neubig}$^{1}$ \quad
\textbf{Chunting Zhou}$^{3}$ \quad
\\
$^1$Carnegie Mellon University\quad
$^2$Instituto Superior Técnico \quad
$^3$Facebook AI Research\\
 \texttt{\{sameerj, vkeshava, smysores\}@cs.cmu.edu}
}

\begin{document}
\maketitle

\begin{abstract}
Evaluation of natural language generation (NLG) is complex and multi-dimensional. Generated text can be evaluated for fluency, coherence, factuality, or any other dimensions of interest. Most frameworks that perform such multi-dimensional evaluation require training on large manually or synthetically generated datasets. In this paper, we study the efficacy of large language models as multi-dimensional evaluators using in-context learning, obviating the need for large training datasets. Our experiments show that in-context learning-based evaluators are competitive with learned evaluation frameworks for the task of text summarization, establishing state-of-the-art on dimensions such as relevance and factual consistency. We then analyze the effects of factors such as the selection and number of in-context examples on performance. Finally, we study the efficacy of in-context learning-based evaluators in evaluating zero-shot summaries written by large language models such as GPT-3. Our code is available at \url{https://github.com/JainSameer06/ICE}
\end{abstract}

\section{Introduction} \label{intro}
Developing comprehensive evaluation frameworks~\cite{deng-2021-compression, yuan-2021-bartscore, zhong-2022-towards} that can evaluate multiple human-interpretable dimensions, 
such as factual consistency~\cite{kryscinski-2020-evaluating, wang-2020-asking} and coherence~\cite{dziri-2019-evaluating, huang-2020-grade}, is important for the advancement of Natural Language Generation (NLG). 
However, similarity-based metrics~\cite{papineni-2002-bleu, lin-2004-rouge, sellam-2020-bleurt, zhao-2019-moverscore, zhang-2020-bertscore} still dominate NLG evaluation in practice. Compared to them, desired multi-dimensional evaluators do not require reference texts for evaluation; and they can easily extend to new explainable evaluation dimensions. 
Recently, \citet{zhong-2022-towards} developed a unified evaluation framework that can generalize to multiple dimensions and text generation tasks.
However, it relies on the construction of synthetic and auxiliary data for the finetuning of a pre-trained language model, requiring in-depth knowledge and significant engineering effort for each dimension. Furthermore, the inclusion of new dimensions requires (continued) training of the model, and might affect the performance on other dimensions in unforeseen ways. 

\begin{figure}
    \centering
    \includegraphics[width=240pt]{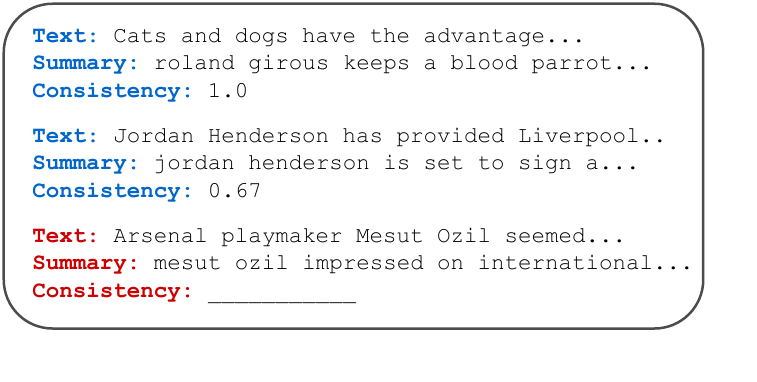}
    \vspace{-10mm}
    \caption{Our prompt design to evaluate the consistency of the summary in red, illustrated using two in-context examples (in blue). To evaluate other aspects, we remove the source text or replace it with a reference.}
    \label{fig:prompt}
    \vspace{-3mm}
\end{figure}

In this work, we propose to use \emph{in-context} learning ~\cite{brown-2020-language} with large language models (LLMs) --- a commonly used method to perform many tasks by utilizing only a few input-output examples ---
to perform multi-dimensional text evaluation in a unified fashion. Compared to pre-trained evaluators that need specialized supervised training for each dimension, our \textbf{I}n-\textbf{C}ontext learning-based \textbf{E}valuator (\textsc{Ice}) framework is:

\begin{itemize*}
    \item Learning-free. It does not require supervised fine-tuning on large annotated (synthetic) training data, requiring only a handful of samples at inference time.
    \item Extensible. To evaluate new dimensions, it does not rely on large amounts of human judgments or the construction of new synthetic data, using only a natural language prompt consisting of a small number of example pairs to ascertain the properties associated with a given quality aspect. 
\end{itemize*}


In this paper, using text summarization as a test bed, we show that with a simple prompt design, \textsc{Ice} is competitive with state-of-the-art trained evaluators on multi-dimensional evaluation of model-produced summaries, establishing a new state-of-the-art on dimensions such as relevance and factual consistency. To study the robustness of the evaluator to the selection of in-context examples, we analyze the factors that affect the performance of \textsc{Ice}, such as the number of in-context examples and sampling procedures when picking in-context examples from a set of candidates.
We find \textsc{Ice} to be robust to the selection of in-context examples and observe a slight improvement in performance as the number of examples is increased. Finally, in light of the recent work~\cite{goyal-news-2022} that points to the misalignment 
of existing evaluation metrics with human preference in evaluating zero-shot summaries generated by LLMs such as GPT-3 \cite{brown-2020-language}, we study the effectiveness of \textsc{Ice} in evaluating zero-shot summaries generated by GPT-3. We find that \textsc{Ice} evaluations agree closely with human judgments on such summaries. 








\section{Methodology}
\subsection{Problem Statement} \label{problem}
Given a sequence $\mathbf{x}$ that is input to an NLG system and a system-generated output sequence $\mathbf{y}$, an evaluation framework outputs a score $s$ that captures the quality of $\mathbf{y}$, either with or without the help of a human-generated reference output $\mathbf{r}$.%
\footnote{Specifically for summarization, most learned frameworks evaluate relevance through reference-based evaluation.}
In case of multi-dimensional evaluation where we are interested in assessing $\mathbf{y}$ over $d$ quality metrics, we instead get a vector $\mathbf{S} = (s_1, s_2, ..., s_d)$  over diverse dimensions (e.g., coherence, fluency). Depending on the dimension, there is sometimes a need to condition an evaluation on $\mathbf{x}$ (such as to evaluate consistency in summarization).
We evaluate our method over four dimensions:
\begin{itemize*}
    \item Consistency: The factual correctness of a summary given the source text.
    \item Relevance: The property of capturing salient information from the source.
    \item Fluency: A measure of the quality of the individual sentences in the summary.
    \item Coherence: A measure of the quality, organization, and structure of sentences in the summary.
\end{itemize*}

\subsection{Prompt Design \& Score Extraction} \label{subsection:prompt}

\textsc{Ice} relies on an LLM (we use the \texttt{text-davinci-003} model of GPT-3) to make predictions. It takes in a prompt that consists of a small number of in-context examples, each of which consists of generated text and its corresponding quality score as a numeric string. The prompt ends with a test example, for which the model predicts a score (Figure \ref{fig:prompt}).

The input contains the model-generated text (summary), in addition to which it might contain additional information such as the source text or references, depending on the dimension. To evaluate \texttt{fluency} and \texttt{coherence}, our prompts use in-context examples consisting of generated summaries and corresponding scores. For \texttt{consistency} and \texttt{relevance}, we use the source text and a reference summary respectively, in addition to the generated summary. We pass this prompt to a GPT-3 model, with sampling temperature set to $0$ to elicit deterministic responses. We parse the model response--decoded numeric string--as the dimension score.


\begin{table*}
\centering
\small
\begin{tabular}{lcccccccc}
\toprule
\textbf{Metric} & \multicolumn{2}{c}{\textbf{Coherence}} & \multicolumn{2}{c}{\textbf{Consistency}} & \multicolumn{2}{c}{\textbf{Fluency}} & \multicolumn{2}{c}{\textbf{Relevance}}\\
 & $\rho$ & $\tau$ & $\rho$ & $\tau$ & $\rho$ & $\tau$ & $\rho$ & $\tau$ \\
\cmidrule(lr){1-1} \cmidrule(lr){2-3}  \cmidrule(lr){4-5} \cmidrule(lr){6-7} \cmidrule(lr){8-9}
CTC  & - & - & 0.425 & 0.340 & - & - & \textbf{0.495} & 0.364\\
BARTScore  & 0.445 & 0.340 & 0.380 & 0.314 & 0.345 & 0.283 & 0.357 & 0.274\\
UniEval   & \textbf{0.591} & \textbf{0.424} & 0.433 & 0.348 & \textbf{0.445} & \textbf{0.349} & 0.473 & 0.343\\
\midrule
\textsc{Ice} (Uniform Sampling)   & 0.476 & 0.388 & \textbf{0.486} & \textbf{0.466} & 0.366 & 0.328 & 0.467 & 0.384\\
\textsc{Ice} (Stratified Sampling)   & 0.497 & 0.387 & 0.298 & 0.263 & 0.397 & 0.348 & 0.485 & \textbf{0.396}\\
\bottomrule
\end{tabular}
\caption{Summary-level Spearman and Kendall-Tau correlations of different metrics on the SummEval benchmark}

\label{tab:main}
\vspace{-3mm}
\end{table*}

\subsection{Selection of In-context Examples} \label{subsection:sampling}


By default, we use 4 in-context examples in our prompts, as this is the largest number that fits within the context window of GPT-3. We experiment with two sampling procedures (Appendix \ref{app:sampling}) to obtain 4 examples from a pool of examples:
\begin{enumerate}[noitemsep]
    \item \textbf{Uniform Random Sampling}. We randomly select 4 summaries from the pool of examples. This causes the examples to follow the same distribution as the example pool.
    \item \textbf{Stratified Sampling}. We bucket the range of scores, i.e. $[0, 1]$, into 4 equal partitions and randomly sample one summary from each one. This causes examples to be representative of the range of scores in the example pool.
\end{enumerate}

We avoid using synthetically generated data~\cite{kryscinski-2020-evaluating, zhong-2022-towards} since the kind of errors made by generation models is often different from the errors present in the negative examples in these datasets~\cite{goyal-2021-annotating}. We instead elect to use (a few) human evaluations of model-generated text in order to make the in-context examples as representative of real errors as possible. We do this by splitting the meta-evaluation dataset and using a partition as an in-context example pool, as described in Section \ref{dataset}.

\section{Experiments} \label{sec:setup}
\subsection{Datasets \& Baselines} \label{dataset}
We use the SummEval dataset \cite{fabbri2020summeval}\footnote{\url{https://github.com/Yale-LILY/SummEval}} to meta-evaluate our evaluation framework. SummEval collects human evaluation annotations for 16 summarization systems on 100 articles sampled from the CNN/DailyMail corpus, for a total of 1600 summary-level annotations. Each summary is evaluated on four dimensions described in Section \ref{subsection:prompt}.

To get a pool of in-context examples, we keep aside a small subset (64 examples) of the SummEval dataset to pick in-context examples from, and use the rest (1536 examples) as the test set for meta-evaluation (evaluating the baselines on this same test set). Further details are in Appendix \ref{app:selection}. 

We compare \textsc{Ice} to the following state-of-the-art multi-dimensional evaluators: (1) \textbf{CTC}~\cite{deng-2021-compression} uses information alignment between generated outputs and references or inputs; (2) \textbf{BARTScore}~\cite{yuan-2021-bartscore} uses the conditional probability of a sequence given inputs or references; and (3) \textbf{UniEval}~\cite{zhong-2022-towards} uses a question-answering framework (e.g. \textit{"Is this a coherent summary?")} to calculate metrics. 

Following~\citet{liu-2021-explainaboard,zhong-2022-towards}, we assess performance by computing summary-level Spearman and Kendall-Tau correlations between predicted scores and human judgements.

\subsection{Results}
As illustrated in Table \ref{tab:main}, \textsc{Ice} is competitive with fine-tuned baselines despite not requiring any finetuning. It achieves state-of-the-art correlation with human judgments for \texttt{relevance} and \texttt{consistency}. We perform pairwise significance tests and observe that \textsc{Ice} (uniform sampling) does better than UniEval on \texttt{consistency} and \texttt{relevance} on Kendall's Tau with a significance level of 0.05 (Appendix \ref{app:signi}). Additionally, the uniform sampling variant of \textsc{Ice} outperforms BARTScore (which also does not require fine-tuning) across dimensions.

Between the two \emph{sampling procedures} for \textsc{Ice}, we observe that stratified sampling works marginally better for all dimensions other than \texttt{consistency}. Since summaries in the SummEval dataset have perfect or near-perfect human scores for \texttt{consistency} (Figure \ref{fig:summeval}), uniform sampling causes in-context examples to also have near-perfect scores. This appears useful for the model to calibrate its scoring when evaluating \texttt{consistency}, leading to better performance. We explore this in greater detail in \S \ref{subsec:samplinganalysis}. While the same reasoning could hold for \texttt{fluency}, we observe both here and in \S\ref{subsection:number} that \texttt{fluency} scores are quite stable. Given that \texttt{fluency} is an easier aspect to evaluate, this stability could be a result of the model possessing a strong notion about \texttt{fluency} from pre-training time that is not modified significantly as the distribution of in-context examples changes~\cite{reynolds-2021-prompt}. Finally, we observe that the performance for \texttt{coherence} and \texttt{relevance} are similar regardless of the sampling procedure. This is because scores for these aspects are spread out in the dataset, which makes uniform and stratified sampling return similar in-context examples.

\section{Analysis}
\label{sec:analyses}
In this section, we analyse the effects of our prompt engineering choices. The comparison between sampling procedures in Section \ref{subsec:samplinganalysis} is performed on the entire test set but the experiments in Sections \ref{subsec:examples} and \ref{subsection:number} are performed on a test set sample of size 200 to control costs. The analyses in Sections \ref{subsec:samplinganalysis} and \ref{subsec:examples} use four in-context examples.
\subsection{Analyzing the Sampling Procedures} \label{subsec:samplinganalysis}
\begin{figure}
    \centering
    \includegraphics[width=225pt]{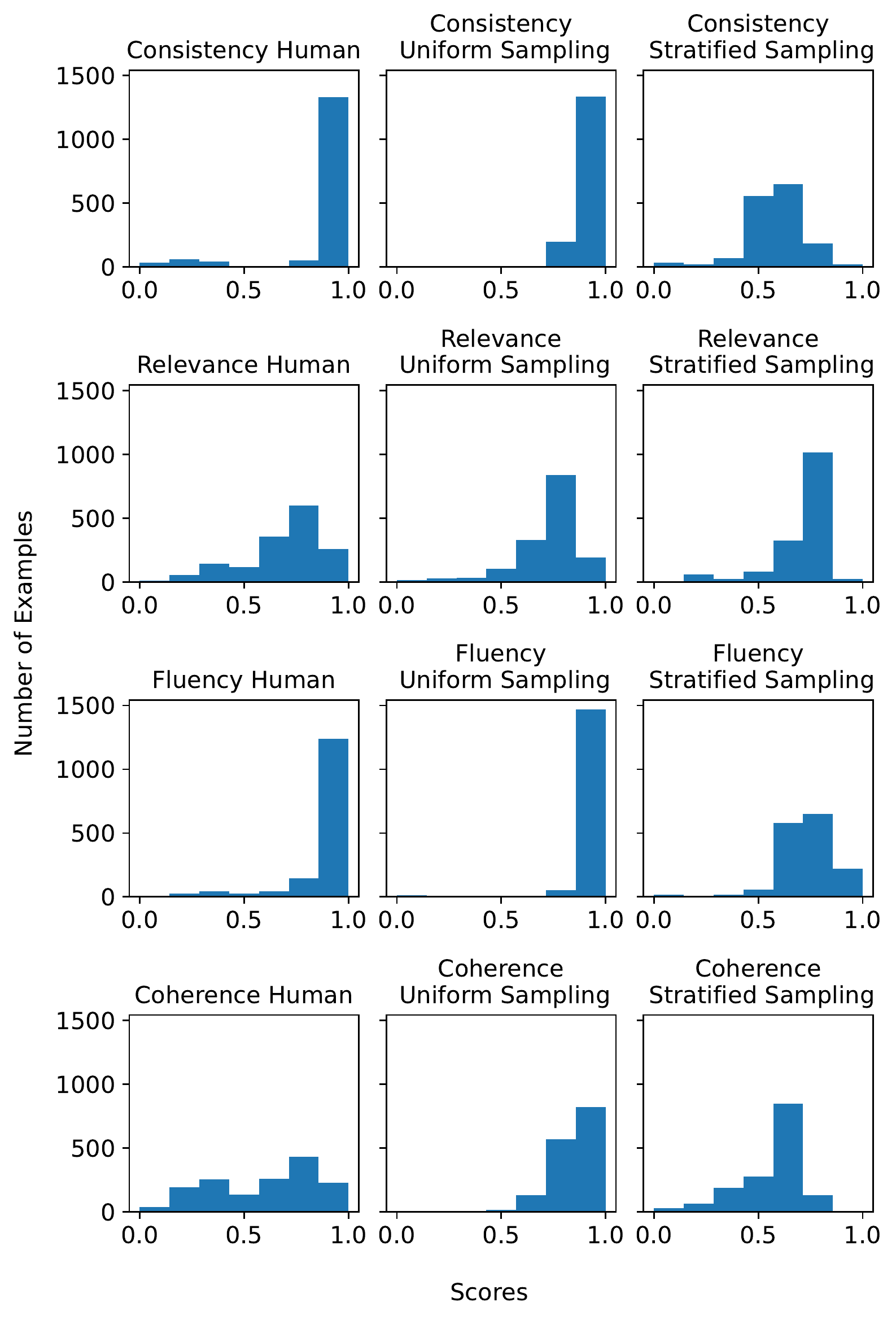}
    \caption{Distributions of human scores and predicted scores using \textsc{ICE} with uniform and stratified sampling on the SummEval benchmark}
    \label{fig:summeval}
    \vspace{-2mm}
\end{figure}

Figure \ref{fig:summeval} illustrates that the prediction distributions from uniform and stratified sampling differ the most when the true distribution is skewed, such as for \texttt{consistency}. In such a case, stratified sampling selects in-context examples from the entire domain regardless of the true distribution. This forces predictions towards a centered distribution, which can cause the performance drop we observe in Table \ref{tab:main} when evaluating \texttt{consistency} using stratified sampling. Uniform sampling, on the other hand, selects examples that represent the true distribution, making model predictions more closely reflect the true distribution. 

A drawback of uniform sampling is sub-optimal calibration in low-probability regions of the true distribution. For instance, if uniform sampling is used to evaluate \texttt{consistency}, the model might not see in-context examples with (say) scores less than 0.3 (Figure \ref{fig:summeval}). This can affect output calibration in that region.
Nonetheless, we suggest using uniform sampling in general. It is more stable and its prediction distribution closely follows the true distribution. For dimensions where it underperforms stratified sampling, the margins are less significant. Finally, even when \textsc{Ice} (uniform sampling) scores are calibrated differently from human scores, they still rank summary-quality correctly, insofar as our main results (Table \ref{tab:main}) show that they compete with state-of-the-art on ranking-based metrics like Kendall-Tau and Spearman correlation. We use uniform sampling to select in-context examples in Sections \ref{subsec:examples} and \ref{subsection:number}.

\subsection{Effect of Selection of In-context Examples} \label{subsec:examples}
\begin{figure}
    \centering
    \includegraphics[width=0.95\columnwidth]{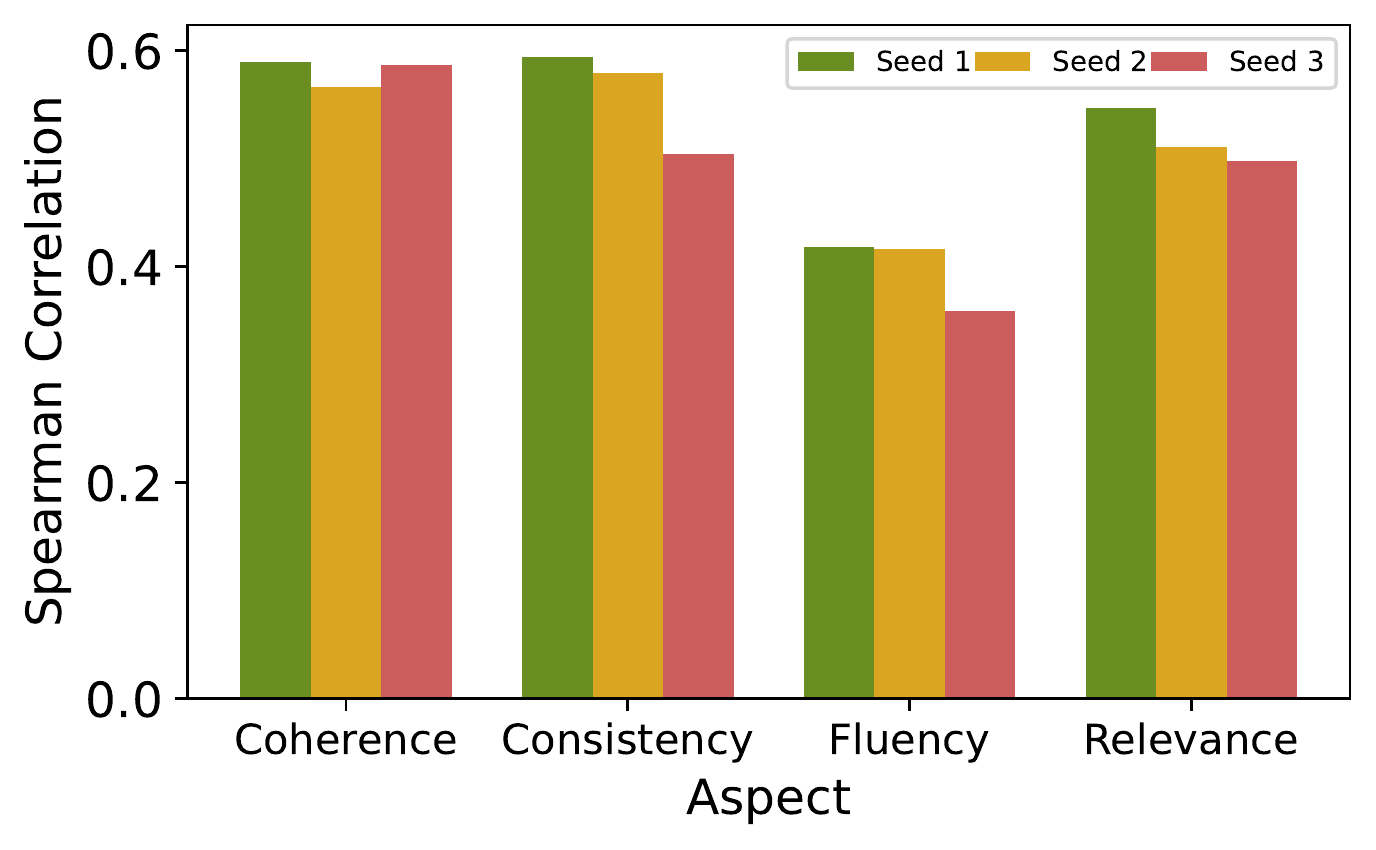}
    \caption{Effect of sampling different in-context examples. The performance over the same test set is observed to be robust to the choice of in-context examples.}
    \label{fig:ice}
\end{figure}

In order to determine whether performance is robust to the choice of in-context examples, we evaluate our test set using three different random sets of in-context examples. We observe in Figure \ref{fig:ice} that for a given dimension, the maximum variation across three seeds is about 7 points, suggesting reasonably stable performance across the choice of in-context examples.

\begin{figure}
    \centering
    \includegraphics[width=0.95\columnwidth]{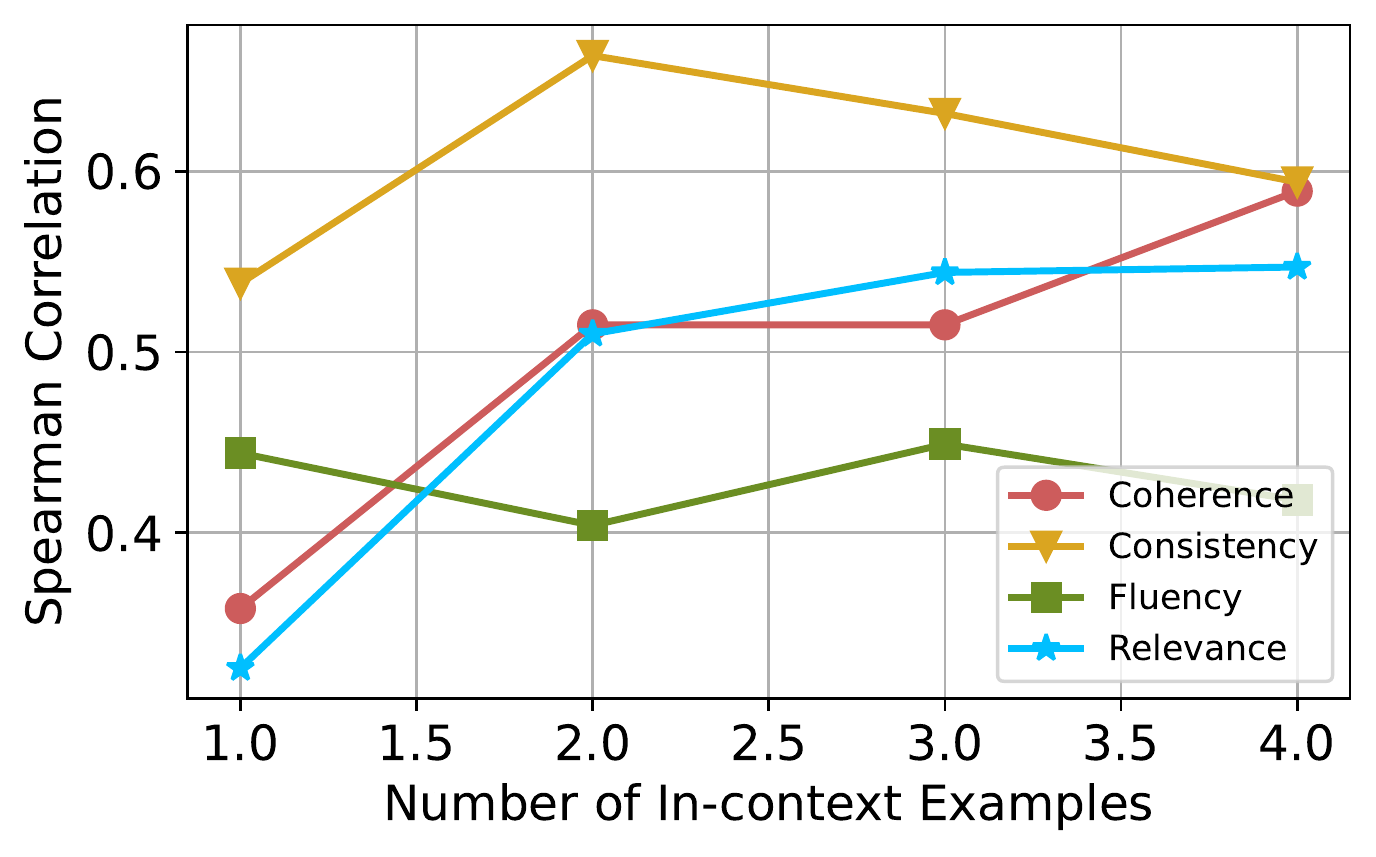}
    \vspace{-2mm}
    \caption{Effect of varying the number of in-context examples.}
    \label{fig:number}
\end{figure}
\subsection{Effect of Number of In-context Examples} \label{subsection:number}
We evaluate our test set using different numbers of in-context examples (Figure \ref{fig:number}). We observe that only for \texttt{relevance} and \texttt{coherence} does performance show improvement as we increase the number of examples. One reason for this could be the distribution of scores for a given dimension in the test set (Figure \ref{fig:summeval}). Concretely, \texttt{consistency} and \texttt{fluency} mostly have near-perfect scores and therefore do not benefit from more samples while the scores for \texttt{coherence} and \texttt{relevance} are spread out and therefore more samples allow representation over the whole range of scores.

Another observation is that even for \texttt{coherence} and \texttt{relevance}, performance with a single in-context example reaches near that achieved by some of the weaker fine-tuned baselines in Table \ref{tab:main}. This suggests that the model possesses the notion of the evaluation task from pre-training itself, which is in line with recent work ~\cite{reynolds-2021-prompt, min-2022-rethinking} that suggests that demonstrations help extract this knowledge.

Finally, we note that calibration can potentially be improved by increasing the number of examples. For instance, we observed that the four in-context examples that the uniform sampling procedure chose for coherence in Figure \ref{fig:summeval} had scores that fall between 0.7 and 1.0. This concentrates the prediction distribution in that range. The probability of such an event will reduce as the number of examples is increased further.

\begin{table}
\centering
\renewcommand{\arraystretch}{0.92}
\resizebox{0.95\columnwidth}{!}{%
\begin{tabular}{c|l|cccc|c}
\toprule
\textbf{Metric} & \textbf{Model} & \textbf{Coh.} & \textbf{Con.} & \textbf{Flu.} & \textbf{Rel.} & \textbf{Overall}\\

\midrule
\multirow{3}{*}{\textit{Human}} & GPT-3   & \cellcolor{orange!20}4.85 & \cellcolor{orange!20}4.73 & \cellcolor{orange!20}4.97 & \cellcolor{orange!20}4.65 & \cellcolor{orange!20}4.80\\

& BRIO   & 4.57 & 4.65 & 4.88 & 4.48 & 4.65\\

&T0    & \cellcolor{blue!20}4.15 & \cellcolor{blue!20}4.47 & \cellcolor{blue!20}4.78 &  \cellcolor{blue!20}3.68 & \cellcolor{blue!20}4.27\\

\midrule
\multirow{3}{*}{ROUGE-L} & GPT-3  & \multicolumn{4}{c|}{\multirow{3}{*}{-}} & \cellcolor{blue!20}22.09\\

& BRIO   & \multicolumn{4}{c|}{}  & \cellcolor{orange!20}28.20\\

& T0    & \multicolumn{4}{c|}{}  & 26.63\\

\midrule
\multirow{3}{*}{BARTSc.} & GPT-3 & \cellcolor{blue!20}-1.25 & \cellcolor{blue!20}-1.25 & \cellcolor{blue!20}-1.25 & \cellcolor{blue!20}-1.25 & \cellcolor{blue!20}-1.25\\

& BRIO  & \cellcolor{orange!20}-0.71 & \cellcolor{orange!20}-0.71 & \cellcolor{orange!20}-0.71 & \cellcolor{orange!20}-0.71 & \cellcolor{orange!20}-0.71\\

& T0 & -0.96 & -0.96 & -0.96 & -0.96 & -0.96\\

\midrule
\multirow{3}{*}{\textsc{Ice}} & GPT-3   & \cellcolor{orange!20}0.908 & \cellcolor{orange!20}0.996 & \cellcolor{orange!20}0.994 & \cellcolor{orange!20}0.849 & \cellcolor{orange!20}0.937\\

& BRIO   & 0.896 & 0.993 & 0.993 & 0.834 & 0.929\\

& T0    & \cellcolor{blue!20}0.890 & \cellcolor{blue!20}0.981 &\cellcolor{blue!20} 0.985 &  \cellcolor{blue!20}0.761 &\cellcolor{blue!20} 0.904\\
\bottomrule
\end{tabular}%
}
\caption{System-level scores from human annotations and automatic metrics. For each aspect, we color a given metric's highest/lowest rated system with orange/purple.}
\label{tab:humananno}
\end{table}

\section{Using \textsc{Ice} to Evaluate Zero-Shot Prompting Models} \label{sec:gptsum}
Recent work by ~\citet{goyal-news-2022} showed that standard reference-based and reference-free metrics are not reliable in evaluating zero-shot summaries written by models such as GPT-3. Through a human study comparing summaries from three systems--GPT-3, BRIO, and T0--they observed that while humans prefer GPT-3 summaries, automatic evaluators consistently score GPT-3 summaries lower than summaries from other models.


We study the efficacy of \textsc{Ice} in evaluating zero-shot summaries written by GPT-3 at a dimension level. We use the set of 500 CNN articles from \citet{goyal-news-2022}, with summaries from GPT-3, BRIO, and T0 for each article. We sample 100 of these articles and have three annotators rate summaries for each of the dimensions defined in Section \ref{subsection:prompt} on a scale of $\{1,2,3,4,5\}$. We use \textsc{Ice}, ROUGE, and BARTScore (all of which do not require training data) to evaluate the summaries and present system-level results in Table \ref{tab:humananno}. \looseness=-1

We observe that \textsc{Ice} agrees with human judgments for each dimension and overall preferences while existing reference-based and reference-free metrics such as ROUGE and BARTScore\footnote{SummEval  annotations are all based on the source, and the src-to-hyp version of BARTScore performs best across dimensions for this benchmark. We use this version for all dimensions, leading to identical scores. We format BARTScore results unlike ROUGE-L because in theory BARTScores can differ across dimensions for an arbitrary benchmark.} consistently rate GPT-3 summaries low. ~\citet{goyal-news-2022} suggest that most existing evaluation metrics reward summaries that imitate references, while GPT-3 summaries are zero-shot and not trained to imitate human-written references, which is likely why they are penalized by most existing evaluators. However, since \textsc{Ice} is not based on reference similarity (except when evaluating \texttt{relevance}) and is also not trained with reference summaries, it is able to better evaluate GPT-3 summaries and agrees with human preferences.

\section{Conclusion}
We show that in-context learning can be used for NLG evaluation as an alternative to fine-tuned evaluation metrics. Using a small number of examples, in-context learning evaluators can reach or exceed the state-of-the-art on multi-dimensional evaluation and that this is robust to the choice of in-context examples. Finally, we show that in-context learning evaluators align well with human judgements when evaluating summaries written by GPT-3.

\section*{Limitations}
While \textsc{Ice} does not require fine-tuning on large amounts of data, it requires querying a powerful LLM at inference time (we use GPT-3 for our experiments which has 175 billion parameters). This can be a pay-per-use model or an open-source model such as BLOOM. This makes a downstream system that uses \textsc{Ice} reliant on an external dependency, which carries the risk of the external dependency failing.

Relatedly, in this paper, we are limited due to monetary constraints in a variety of experiments we perform. For instance, we restrict ourselves to text summarization and use samples of benchmark meta-evaluation suites during some of our experiments. We leave the investigation of using \textsc{Ice} for other dimensions and downstream tasks for future work.
\bibliography{anthology,custom}

\begin{thebibliography}{19}
\expandafter\ifx\csname natexlab\endcsname\relax\def\natexlab#1{#1}\fi

\bibitem[{Brown et~al.(2020)Brown, Mann, Ryder, Subbiah, Kaplan, Dhariwal,
  Neelakantan, Shyam, Sastry, Askell, Agarwal, Herbert-Voss, Krueger, Henighan,
  Child, Ramesh, Ziegler, Wu, Winter, Hesse, Chen, Sigler, Litwin, Gray, Chess,
  Clark, Berner, McCandlish, Radford, Sutskever, and
  Amodei}]{brown-2020-language}
Tom~B. Brown, Benjamin Mann, Nick Ryder, Melanie Subbiah, Jared Kaplan,
  Prafulla Dhariwal, Arvind Neelakantan, Pranav Shyam, Girish Sastry, Amanda
  Askell, Sandhini Agarwal, Ariel Herbert-Voss, Gretchen Krueger, Tom Henighan,
  Rewon Child, Aditya Ramesh, Daniel~M. Ziegler, Jeffrey Wu, Clemens Winter,
  Christopher Hesse, Mark Chen, Eric Sigler, Mateusz Litwin, Scott Gray,
  Benjamin Chess, Jack Clark, Christopher Berner, Sam McCandlish, Alec Radford,
  Ilya Sutskever, and Dario Amodei. 2020.
\newblock \href {https://doi.org/10.48550/ARXIV.2005.14165} {Language models
  are few-shot learners}.

\bibitem[{Deng et~al.(2021)Deng, Tan, Liu, Xing, and
  Hu}]{deng-2021-compression}
Mingkai Deng, Bowen Tan, Zhengzhong Liu, Eric Xing, and Zhiting Hu. 2021.
\newblock \href {https://doi.org/10.18653/v1/2021.emnlp-main.599} {Compression,
  transduction, and creation: A unified framework for evaluating natural
  language generation}.
\newblock In \emph{Proceedings of the 2021 Conference on Empirical Methods in
  Natural Language Processing}, pages 7580--7605, Online and Punta Cana,
  Dominican Republic. Association for Computational Linguistics.

\bibitem[{Dziri et~al.(2019)Dziri, Kamalloo, Mathewson, and
  Zaiane}]{dziri-2019-evaluating}
Nouha Dziri, Ehsan Kamalloo, Kory Mathewson, and Osmar Zaiane. 2019.
\newblock \href {https://doi.org/10.18653/v1/N19-1381} {Evaluating coherence in
  dialogue systems using entailment}.
\newblock In \emph{Proceedings of the 2019 Conference of the North {A}merican
  Chapter of the Association for Computational Linguistics: Human Language
  Technologies, Volume 1 (Long and Short Papers)}, pages 3806--3812,
  Minneapolis, Minnesota. Association for Computational Linguistics.

\bibitem[{Fabbri et~al.(2020)Fabbri, Kry{\'s}ci{\'n}ski, McCann, Xiong, Socher,
  and Radev}]{fabbri2020summeval}
Alexander~R Fabbri, Wojciech Kry{\'s}ci{\'n}ski, Bryan McCann, Caiming Xiong,
  Richard Socher, and Dragomir Radev. 2020.
\newblock Summeval: Re-evaluating summarization evaluation.
\newblock \emph{arXiv preprint arXiv:2007.12626}.

\bibitem[{Goyal and Durrett(2021)}]{goyal-2021-annotating}
Tanya Goyal and Greg Durrett. 2021.
\newblock \href {https://doi.org/10.18653/v1/2021.naacl-main.114} {Annotating
  and modeling fine-grained factuality in summarization}.
\newblock In \emph{Proceedings of the 2021 Conference of the North American
  Chapter of the Association for Computational Linguistics: Human Language
  Technologies}, pages 1449--1462, Online. Association for Computational
  Linguistics.

\bibitem[{Goyal et~al.(2022)Goyal, Li, and Durrett}]{goyal-news-2022}
Tanya Goyal, Junyi~Jessy Li, and Greg Durrett. 2022.
\newblock \href {https://doi.org/10.48550/ARXIV.2209.12356} {News summarization
  and evaluation in the era of gpt-3}.

\bibitem[{Huang et~al.(2020)Huang, Ye, Qin, Lin, and Liang}]{huang-2020-grade}
Lishan Huang, Zheng Ye, Jinghui Qin, Liang Lin, and Xiaodan Liang. 2020.
\newblock \href {https://doi.org/10.18653/v1/2020.emnlp-main.742} {{GRADE}:
  Automatic graph-enhanced coherence metric for evaluating open-domain dialogue
  systems}.
\newblock In \emph{Proceedings of the 2020 Conference on Empirical Methods in
  Natural Language Processing (EMNLP)}, pages 9230--9240, Online. Association
  for Computational Linguistics.

\bibitem[{Kryscinski et~al.(2020)Kryscinski, McCann, Xiong, and
  Socher}]{kryscinski-2020-evaluating}
Wojciech Kryscinski, Bryan McCann, Caiming Xiong, and Richard Socher. 2020.
\newblock \href {https://doi.org/10.18653/v1/2020.emnlp-main.750} {Evaluating
  the factual consistency of abstractive text summarization}.
\newblock In \emph{Proceedings of the 2020 Conference on Empirical Methods in
  Natural Language Processing (EMNLP)}, pages 9332--9346, Online. Association
  for Computational Linguistics.

\bibitem[{Lin(2004)}]{lin-2004-rouge}
Chin-Yew Lin. 2004.
\newblock \href {https://aclanthology.org/W04-1013} {{ROUGE}: A package for
  automatic evaluation of summaries}.
\newblock In \emph{Text Summarization Branches Out}, pages 74--81, Barcelona,
  Spain. Association for Computational Linguistics.

\bibitem[{Liu et~al.(2021)Liu, Fu, Xiao, Yuan, Chang, Dai, Liu, Ye, and
  Neubig}]{liu-2021-explainaboard}
Pengfei Liu, Jinlan Fu, Yang Xiao, Weizhe Yuan, Shuaichen Chang, Junqi Dai,
  Yixin Liu, Zihuiwen Ye, and Graham Neubig. 2021.
\newblock \href {https://doi.org/10.18653/v1/2021.acl-demo.34}
  {{E}xplaina{B}oard: An explainable leaderboard for {NLP}}.
\newblock In \emph{Proceedings of the 59th Annual Meeting of the Association
  for Computational Linguistics and the 11th International Joint Conference on
  Natural Language Processing: System Demonstrations}, pages 280--289, Online.
  Association for Computational Linguistics.

\bibitem[{Min et~al.(2022)Min, Lyu, Holtzman, Artetxe, Lewis, Hajishirzi, and
  Zettlemoyer}]{min-2022-rethinking}
Sewon Min, Xinxi Lyu, Ari Holtzman, Mikel Artetxe, Mike Lewis, Hannaneh
  Hajishirzi, and Luke Zettlemoyer. 2022.
\newblock \href {https://doi.org/10.48550/ARXIV.2202.12837} {Rethinking the
  role of demonstrations: What makes in-context learning work?}

\bibitem[{Papineni et~al.(2002)Papineni, Roukos, Ward, and
  Zhu}]{papineni-2002-bleu}
Kishore Papineni, Salim Roukos, Todd Ward, and Wei-Jing Zhu. 2002.
\newblock \href {https://doi.org/10.3115/1073083.1073135} {Bleu: A method for
  automatic evaluation of machine translation}.
\newblock In \emph{Proceedings of the 40th Annual Meeting on Association for
  Computational Linguistics}, ACL '02, page 311–318, USA. Association for
  Computational Linguistics.

\bibitem[{Reynolds and McDonell(2021)}]{reynolds-2021-prompt}
Laria Reynolds and Kyle McDonell. 2021.
\newblock \href {https://doi.org/10.48550/ARXIV.2102.07350} {Prompt programming
  for large language models: Beyond the few-shot paradigm}.

\bibitem[{Sellam et~al.(2020)Sellam, Das, and Parikh}]{sellam-2020-bleurt}
Thibault Sellam, Dipanjan Das, and Ankur Parikh. 2020.
\newblock \href {https://doi.org/10.18653/v1/2020.acl-main.704} {{BLEURT}:
  Learning robust metrics for text generation}.
\newblock In \emph{Proceedings of the 58th Annual Meeting of the Association
  for Computational Linguistics}, pages 7881--7892, Online. Association for
  Computational Linguistics.

\bibitem[{Wang et~al.(2020)Wang, Cho, and Lewis}]{wang-2020-asking}
Alex Wang, Kyunghyun Cho, and Mike Lewis. 2020.
\newblock \href {https://doi.org/10.18653/v1/2020.acl-main.450} {Asking and
  answering questions to evaluate the factual consistency of summaries}.
\newblock In \emph{Proceedings of the 58th Annual Meeting of the Association
  for Computational Linguistics}, pages 5008--5020, Online. Association for
  Computational Linguistics.

\bibitem[{Yuan et~al.(2021)Yuan, Neubig, and Liu}]{yuan-2021-bartscore}
Weizhe Yuan, Graham Neubig, and Pengfei Liu. 2021.
\newblock \href
  {https://proceedings.neurips.cc/paper/2021/file/e4d2b6e6fdeca3e60e0f1a62fee3d9dd-Paper.pdf}
  {Bartscore: Evaluating generated text as text generation}.
\newblock In \emph{Advances in Neural Information Processing Systems},
  volume~34, pages 27263--27277. Curran Associates, Inc.

\bibitem[{Zhang et~al.(2020)Zhang, Kishore, Wu*, Weinberger, and
  Artzi}]{zhang-2020-bertscore}
Tianyi Zhang, Varsha Kishore, Felix Wu*, Kilian~Q. Weinberger, and Yoav Artzi.
  2020.
\newblock \href {https://openreview.net/forum?id=SkeHuCVFDr} {Bertscore:
  Evaluating text generation with bert}.
\newblock In \emph{International Conference on Learning Representations}.

\bibitem[{Zhao et~al.(2019)Zhao, Peyrard, Liu, Gao, Meyer, and
  Eger}]{zhao-2019-moverscore}
Wei Zhao, Maxime Peyrard, Fei Liu, Yang Gao, Christian~M. Meyer, and Steffen
  Eger. 2019.
\newblock \href {https://doi.org/10.18653/v1/D19-1053} {{M}over{S}core: Text
  generation evaluating with contextualized embeddings and earth mover
  distance}.
\newblock In \emph{Proceedings of the 2019 Conference on Empirical Methods in
  Natural Language Processing and the 9th International Joint Conference on
  Natural Language Processing (EMNLP-IJCNLP)}, pages 563--578, Hong Kong,
  China. Association for Computational Linguistics.

\bibitem[{Zhong et~al.(2022)Zhong, Liu, Yin, Mao, Jiao, Liu, Zhu, Ji, and
  Han}]{zhong-2022-towards}
Ming Zhong, Yang Liu, Da~Yin, Yuning Mao, Yizhu Jiao, Pengfei Liu, Chenguang
  Zhu, Heng Ji, and Jiawei Han. 2022.
\newblock \href {https://doi.org/10.48550/ARXIV.2210.07197} {Towards a unified
  multi-dimensional evaluator for text generation}.

\end{thebibliography}
\bibliographystyle{acl_natbib}

\appendix
\section{Splitting SummEval and the Selection of In-context Examples} \label{app:selection}

We randomly select 4 articles from the SummEval dataset and pick one system-generated summary from each article as an in-context example using the procedures outlined in Section \ref{subsection:sampling}. In other words, we pick $n = 4$ in Figure \ref{fig:prompt}. For a given value of $n$, prompts for evaluating \texttt{consistency} are the longest since they contain entire source articles. We pick $n$ such that \texttt{consistency} prompts fit within the context window of the model. We study the effect of the choice of $n$ in Section \ref{subsection:number}. 

To ensure that there is no overlap in the source article of any in-context example with the source article of any test example, we remove all summaries corresponding to the 4 selected source texts and use the remaining 1536 examples from SummEval as our test set. We ensure the absence of overlap throughout all experiments in Sections \ref{sec:setup}, \ref{sec:analyses}, and \ref{sec:gptsum}.

\section{Sampling Procedures} \label{app:sampling}
\subsubsection{Uniform Random Sampling} \label{uniform}
One summary is picked uniformly at random from the set of 16 summaries for a given source text. We do this for each of the 4 source texts selected to pick in-context examples from. Each of the 4 sampled in-context examples then consists of the selected summary, its human evaluation score on the current aspect of interest, and (optionally) the source text or the reference text.

\subsubsection{Stratified Sampling} \label{stratified} 
Let $A$ denote the score of a summary on the aspect we are evaluating for; then $A \in [0, 1]$. In stratified sampling, we define 4 buckets by the ranges $\{[0, 0.25], (0.25, 0.5], (0.5, 0.75], (0.75, 1.0]\}$. We assign summary $s$ to one of the buckets depending on the value of $A_s$. We do this for each of the 64 in-context example candidate summaries. Finally, we pick 4 summaries from the 64 candidates such that each summary falls into a different bucket and also comes from a different source text. We perform an exhaustive search for such an assignment, and in case no such assignment is possible (this can happen if none of the 64 summaries fall in a given bucket), we pick an arbitrary summary from a randomly selected bucket, ensuring that all 4 summaries come from different source articles.

For both uniform and random sampling, we ensure that each summary corresponds to a different source article. 

\section{Annotation Procedure for Rating GPT-3, BRIO, and T0 Summaries}
Summaries are annotated on a scale of $\{1,2,3,4,5\}$ for \texttt{coherence}, \texttt{consistency}, \texttt{fluency}, and \texttt{relevance} using the annotation instructions from ~\citet{fabbri2020summeval}. 

\section{Use of Existing Evaluation Packages}
We use existing packages for all our baselines--ROUGE, BARTScore, CTC, and UniEval. For ROUGE, we use the native python implementation and report ROUGE-L scores for our experiment in Section \ref{sec:gptsum}. For BARTScore, we use the implementation accompanying the paper with the \texttt{source} to \texttt{hypothesis} setting across all dimensions, as that gives the best correlations with human judgments across dimensions. For UniEval, we use pre-trained model released by the authors to obtain results in Table \ref{tab:main} on the test set of size 1536.

\section{Significance Tests} \label{app:signi}
Since \textsc{Ice} scores for some dimensions are close to UniEval scores, we perform pairwise tests to determine when one method is better than the other. Concretely, we compare performance on 1000 bootstrap samples by randomly selecting 80\% of the test set for each sample. We observe that when using Kendall's Tau, \textsc{Ice} with uniform sampling outperforms UniEval with a significance level of 0.05 on both \texttt{consistency} and \texttt{relevance}. When using Spearman's rank correlation, \text{Ice} again outperforms UniEval on \texttt{consistency}, but the test is inconclusive at that significance level for \texttt{relevance}.

\end{document}